\documentclass[conference]{IEEEtran}
\IEEEoverridecommandlockouts
\usepackage{cite}
\usepackage{amsmath,amssymb,amsfonts}
\usepackage{algorithmic}
\usepackage{graphicx}
\usepackage{textcomp}
\usepackage{xcolor}

\usepackage{caption}
\usepackage{float}

\usepackage{hyperref}
\usepackage{booktabs}
\usepackage{subcaption}

\usepackage[ruled,linesnumbered]{algorithm2e}

\def\BibTeX{{\rm B\kern-.05em{\sc i\kern-.025em b}\kern-.08em
    T\kern-.1667em\lower.7ex\hbox{E}\kern-.125emX}}

\begin{document}

\title{Trust and Resilience in Federated Learning Through Smart Contracts Enabled Decentralized Systems}
\author{
\IEEEauthorblockN{
    Lorenzo Cassano\IEEEauthorrefmark{1},
    Jacopo D'Abramo\IEEEauthorrefmark{1},
    Siraj Munir\IEEEauthorrefmark{2}
    Stefano Ferretti\IEEEauthorrefmark{2}\IEEEauthorrefmark{1}
}
\IEEEauthorblockA{\IEEEauthorrefmark{1}Department of Computer Science and Engineering, University of Bologna, Italy}
\IEEEauthorblockA{\IEEEauthorrefmark{2}Department of Pure and Applied Sciences, University of Urbino Carlo Bo, Italy}
{\tt\small \{lorenzo.cassano2,jacopo.dabramo\}@studio.unibo.it,} \\
{\tt\small s.munir@campus.uniurb.it,stefano.ferretti@uniurb.it}
}

\maketitle

\begin{abstract}
In this paper, we present a study of a Federated Learning (FL) system, based on the use of decentralized architectures to ensure trust and increase reliability. The system is based on the idea that the FL collaborators upload the (ciphered) model parameters on the Inter-Planetary File System (IPFS) and interact with a dedicated smart contract to track their behavior. Thank to this smart contract, the phases of parameter updates are managed efficiently, thereby strengthening data security.
We have carried out an experimental study that exploits two different methods of weight aggregation, i.e., a classic averaging scheme and a federated proximal aggregation. The results confirm the feasibility of the proposal.
\end{abstract}

\begin{IEEEkeywords}
Federated Learning, Blockchain, Decentralized Systems, Machine Learning, Smart Contracts
\end{IEEEkeywords}


\section{Introduction}
Federated Learning (FL) is a Machine Learning (ML) framework that enables multiple parties to collaboratively train a shared model without directly sharing their individual data \cite{Wen2023}. FL holds the promise of enhancing data-driven learning models while safeguarding data owners' privacy. This makes it an appealing approach for developing ML models in several application domains. A prominent example is clinical diagnosis, where patient data privacy is critical, yet data aggregation from diverse sources is necessary.
However, a significant challenge revolves around ensuring FL collaborators actively participate in the protocol while securely contributing their data. 
To this extent, recently several studies have explored the integration of blockchain technology within FL systems \cite{Behera2021FederatedLU,qammar2023securing,10.1145/3524104,8905038,imboccioli2024decentralization}. While existing research often emphasizes on participation incentives, traceability, and security aspects, there is an under-explored dimension that deserves attention, i.e., the impact of delays, number of collaborators, and collaborator failures on system performance and model training accuracy.

In our work, we investigate the performance of our FL system and the accuracy of ML training when one or more collaborators experience failures. 

We propose a decentralized FL system that combines FL with two powerful components: the Inter-Planetary File System (IPFS) and smart contracts (executed over a permissioned Ethereum-like blockchain). Together, they create a tamper-proof storage mechanism for sharing encrypted model parameters. IPFS provides decentralized and fault-tolerant data storage, while smart contracts enforce rules and streamline interactions among participants \cite{Zichichi2020}.
The presence of a smart contract forces Collaborators to adhere to a protocol organized into distinct phases, ensuring orderly updates of model parameters. Additionally, the smart contract automates compliance checks, maintaining the integrity of the FL process \cite{imboccioli2024decentralization}.

In the paper, we present the overall decentralized system and we perform an experimental evaluation under varying failure scenarios among Collaborators. As concerns the FL model assessment, two weight update methods are used, \emph{FedAvg} and \emph{FedProx} \cite{LiSZSTS20}. 
We investigate the FL system’s performance in terms of classification accuracy, using a dataset coming from the healthcare image classification domain, i.e., brain tumor image dataset.
Furthermore, we measure the gas consumption of the smart contract, a critical aspect in blockchain-based FL systems.
Finally, we also evaluate the performance of the data retrieval and update, by both FL Manager and Collaborators, from/to IPFS.
Results confirm the viability of using decentralized systems in Federated Learning.

The remainder of this paper is organized as follows. 
Section \ref{sec:archi} describes the proposed framework. 
Section \ref{sec:perf} describes the methodology, while 
Section \ref{sec:res} outlines the experimental results. 
Section \ref{sec:conc} provides some concluding remarks.

\section{System Architecture}\label{sec:archi}

Our system architecture comprises four key actors. The first two are the classic actors involved in a typical FL system, while the last two ones are those of typical decentralized systems \cite{Zichichi2020}:
\begin{itemize}
\item \textbf{FL Manager}: The Manager acts as the aggregator for the classification model parameters used in FL. It triggers requests, collects ML parameters from other Collaborators, and monitors parameter updates.
\item \textbf{FL Collaborators}: these are the nodes participating to FL model training. They retrieve data, locally train their models, and upload the trained parameters to the system.
\item \textbf{Permissioned Blockchain}: This blockchain executes a Federated Learning Smart Contract (FLSC). The FLSC autonomously regulates parameter exchanges among participants, ensuring adherence to the protocol. Additionally, the use of blockchain digests prevents tampering with parameters, enhancing security \cite{Bigini202297,Serena2022}.
\item \textbf{Decentralized File Storage}: The Inter-Planetary File System (IPFS) stores the ML model parameters.
\end{itemize}


\subsection{System Interaction Overview}





\begin{figure}[h]
    \centering
    \includegraphics[width=\linewidth]{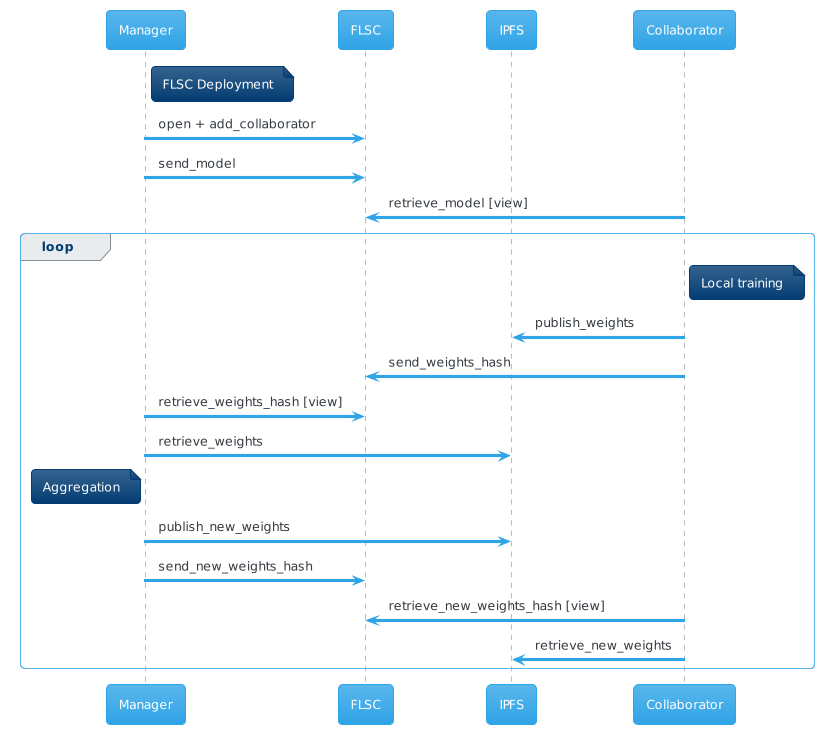}
    \caption{System diagram}\label{fig:diagram}
\end{figure}

The diagram of the interactions among system components is shown in Figure \ref{fig:diagram}.
The protocol works as follows:
\begin{enumerate}
    \item The Manager deploys and initiates the Federated Learning Smart Contract (FLSC), inviting multiple Collaborators (for simplicity, only one is depicted in the figure).
    \item The Manager publishes the model details required for compilation within FLSC.
    \item Collaborators retrieve and adopt the model from the FL Smart Contract.
\end{enumerate}
\noindent\textbf{FL Loop}
\begin{enumerate}
    \setcounter{enumi}{3}
    \item Each Collaborator trains its local model using its dataset. By keeping data decentralized, FL avoids the need for centralizing sensitive information. This brings positive aspects possibly related to enhanced privacy, data protection, regulatory compliance, and scalability.
    \item Periodically, Collaborators publish updated local parameters on IPFS and send their parameter hashes (along with retrieval information) to FLSC. Parameters are encrypted sequentially using the Collaborator's private key and the Manager's public key. This ensures confidentiality and verifiability — only the Manager can read the data, and their origin is guaranteed.
    \item The Manager decrypts these parameters using its private key and the Collaborator's public key. The Manager further validates parameters by comparing their digest to the hashed value stored in the smart contract.
    \item Based on an aggregation function, the Manager updates the ML model parameters, considering results obtained during training.
    \item The Manager securely uploads the updated ML model parameters to IPFS, using the same encryption approach as previously described. These parameters represent the refined knowledge gained during the training process.
    \item The Manager sends the digests (hashes) of these parameters to the smart contract. These digests serve as verifiable references for the stored parameters.
    \item Collaborators retrieve the updated model parameters from IPFS. They decrypt the parameters using their private keys. Collaborators then verify the validity of these parameters through the FL Smart Contract, following the same process as before. This validation ensures that the parameters have not been tampered with and align with the agreed-upon protocol. Once validated, Collaborators incorporate these updated model parameters into their local training during the subsequent iteration.
\end{enumerate}

In our current implementation, Collaborators receive no rewards for parameter updates, assuming protocol compliance. However, our framework allows for alternative security-enhancing or incentive-based approaches. For example, participants might be automatically rewarded based on their contributions \cite{Behera2021FederatedLU,qammar2023securing,zichichi2022data,Barbàra2023166}. 
\subsection{The Federated Learning Smart Contract}
The FLSC has been implemented in Solidity. 
The contract structure incorporates methods that align with the state phases depicted in Figure \ref{fig:diagram}, i.e,~all messages involving the FLSC. It establishes four states for the FL process: OPEN, START, LEARNING, and CLOSE. These states guide the progression of the FL process, ensuring it adheres to predefined rules. This structured approach forces Collaborators to engage with the FLSC only when necessary. Furthermore, the phase definitions facilitate synchronization among all parties involved in FL, effectively addressing issues related to node failures. This implies that if a node fails or is late in transmitting the hash of its weights, the training process is not hindered, and weight updates proceed without considering the contribution of that particular node.

Collaborators are added to the contract during the initialization phase, allowing only authorized users to participate. Collaborators can be added by the contract owner, and their interactions are monitored throughout the FL process.
In Figure \ref{fig:diagram}, we highlight the view functions, which correspond to methods within the smart contract that do not alter the contract state. These functions do not involve gas consumption for their execution and provide read-only access to specific data or parameters.
The contract includes security measures to restrict unauthorized access and ensure that actions are carried out only by authorized users. During each round of the FL, Collaborators can perform specific actions only once, and the contract owner (Manager) has control over the FL process.
Finally, the contract emits events to signal state transitions during the FL process, providing transparency and auditability.

\subsection{On the Security and Reliability of the Protocol}
The protocol is designed to handle the possibility of Collaborators failures. However, it clearly assumes that the Manager cannot fail. The Manager is responsible for coordinating all activities, including sending messages to the FLSC, computing the updated weights, uploading the new values to IPFS, and notifying the Collaborators via smart contract calls, which in turn will emit events that the Collaborators can wait for.

The security of the protocol also relies on the Manager. Collaborators encrypt their weights using the Manager's public key and their own private key. This ensures verifiability (i.e., anyone can verify that the message was generated by a specific Collaborator) and confidentiality (i.e., only the Manager can decrypt the values).

These aspects make the Manager a possible single point of failure for the system. If the Manager fails, no one can carry out the training phase, as no one can invoke the appropriate functions of the FLSC smart contract or decrypt the values.

This limitation can be addressed by adopting one of the classic consensus protocols in an asynchronous network for managing a primary node election \cite{10.1145/322186.322188}. Any problems related to data encryption could be solved by using mechanisms that facilitate the use of encrypted decentralized file storage while enabling data sharing, such as key re-distribution techniques and multi-party computation \cite{Barbàra2023166}.

\section{Experimental Evaluation}\label{sec:perf} 

We performed an experimental analysis to evaluate, on one side, the performance of the FL system, based on different aggregation methods and system setting. We also varied the amount of total nodes involved in the FL, with the possibility of having node failures during the training process, so as to assess how these node failures impact the overall performance. On the other hand, we measured the gas fees of the smart contract system, based on the amount of involved nodes. 

\subsection{The Federated Learning Model and Weight Update Methods}
Collaborators used the same ML approach for classification, i.e., a three-layered, two-dimensional Convolutional Neural Network (CNN) model, coupled with a final fully connected layer at the end for the final classification of each image.
The models are trained with the same setup: an Adam optimizer, a learning rate set to 10e-3, batch size equal to 32, softmax as the last activation function.

In this work, we used two specific FL weight update methods, inspired by \cite{LiSZSTS20}. In few words, we will denote with \emph{FedAvg} the classic approach that assumes that all Collaborators contribute equally and that the dataset is evenly distributed across all nodes. Thus, when weights need to be updated at the end of a loop iteration, the novel weight is just the average value of weights received from the Collaborators. 
In this case, the classic cross-entropy is used as the loss function.

Federated Proximal (\emph{FexProx}), instead, removes the uniformity assumption.
Proximal optimization involves adding a regularization term to the objective function to encourage specific properties (e.g., sparsity).
At the end of the FL iteration happened at timestep $t$, weights are updated locally in order to minimize a loss function of this form
$$\arg \min\limits_w  F(w) + \frac{\mu}{2} \|w - w^t\|_2^2,$$
where $F(w)$ represents the local loss function, $w$ denotes the local weights to be identified, and $w^t$ signifies the global weights sent by the Manager at the beginning of timestep $t$. 
The Manager collects these weights coming from Collaborators, and then computes a novel version of these weights $w^{t+1}$ by taking the average of the received values.

During our experiments, we tried with different values for the $\mu$ hyper-parameter, finding that after the tuning the best results were obtained with $\mu=0.001$. Thus, we show results obtained according to this setting.

\subsection{Brain Tumor Dataset}
The dataset we employed during the tests is Brain Tumor MRI Dataset: https://www.kaggle.com/datasets/iashiqul/brain-tumor-mri-image-classification. 
This dataset contains 7,022 human brain MRI images classified into four classes: glioma, meningioma, no tumor, and pituitary.

\section{Results}\label{sec:res} 

The implemented system involves various technologies, i.e., i) FL techniques for the decentralized analysis of datasets, which are partitioned across multiple sources; ii) blockchain technologies, particularly the FLSC, to coordinate the training phases, iii) IPFS, used for the decentralized yet secure exchange of information, thanks to data encryption. To evaluate this type of system, it is thus necessary to consider its various aspects.
In this section, we show results obtained by using the FL system on the mentioned dataset related to image classification in healthcare, which is a typical use-case application for FL. Then, we will show the performance of the FLSC used to orchestrate the interaction among FL nodes \cite{COELHO2023113,10.1145/3501813,BRISIMI201859}. Finally, we show also experimental results obtained for the weights retrieval and transmission to IPFS.

\subsection{FL performance}
Table \ref{tab:brain_best} shows the averaged accuracy and F1 score obtained by the two FL aggregation techniques, against a centralized approach, i.e., the one that uses a classic ML, where the whole dataset is stored at a central node. In this case, er show results obtained in the best configuration, i.e., when the amount of employed Collaborators was equal to 5, with no failing nodes, together with the Manager (i.e., 6 nodes in total). Without any surprise, best results are obtained with the centralized approach. Having all the instances in the same data-store eases the training, and this leads to best results. However, we discussed already that in certain scenarios this is not possible. Thus, the centralized approach should be taken as an upper bound for the FL methods. In this respect, it is interesting to observe that the FL schemes perform quite well, especially FedAvg that is only 0.02 scores below centralized.

\begin{table}[h]
\centering
\caption{Accuracy Results. Number of Collaborators = 5}\label{tab:brain_best}
\begin{tabular}{|l|l|l|l|}
\hline
\textbf{Agg. Method} & \textbf{Accuracy} & \textbf{F1\_score} \\
\hline
Centralized & 0.98 & 0.98 \\
\hline
FedAvg  &  0.96   &   0.96    \\
\hline
FedProx &  0.96 & 0.95 \\
\hline
\end{tabular}
\end{table}

\begin{table}[h]
\centering
\caption{Classification results for FedAvg and FedProx. Number of Collaborators = 5. Accuracy FedAvg=0.96, FedProx=0.96}\label{tab:BT_combined_best}
\begin{tabular}{|l|l|l|l|l|}
\hline
& \textbf{Precision} & \textbf{Recall} & \textbf{F1-score} & \textbf{Support} \\
\hline
\textbf{Glioma} & 0.99 / 0.94 & 0.90 / 0.94 & 0.94 / 0.94 & 299 \\
\textbf{Meningioma} & 0.89 / 0.92 & 0.95 / 0.90 & 0.92 / 0.91 & 301 \\
\textbf{Notumor} & 0.98 / 0.98 & 0.99 / 0.99 & 0.99 / 0.98 & 381 \\
\textbf{Pituitary} & 0.98 / 0.98 & 0.99 / 0.99 & 0.99 / 0.99 & 300 \\
\hline
\end{tabular}
\end{table}

Table \ref{tab:BT_combined_best} shows FL results related to each class.
In each cell, the first value corresponds to FedAvg and the second value corresponds to FedProx. 
Both methods seem to perform well, with FedAvg which is slightly superior to FedProx in terms of F1 score.

\begin{figure*}[h]
    \centering
    \begin{subfigure}[b]{0.45\textwidth}
        \centering
        \includegraphics[width=.9\textwidth]{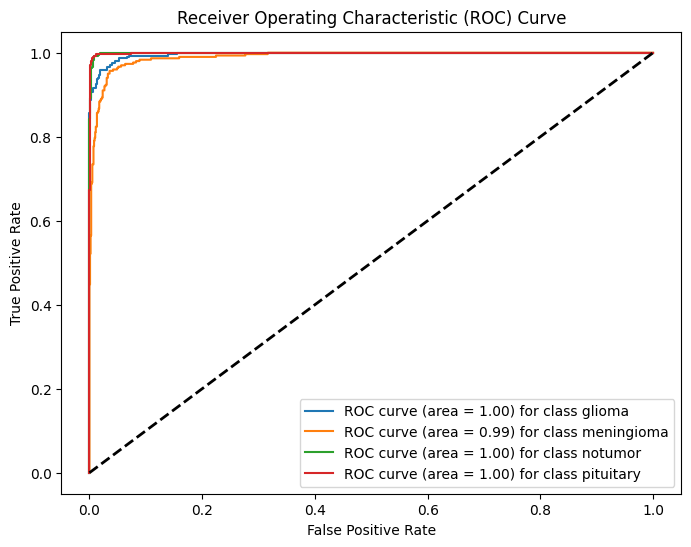}
        \caption{FedAvg}
    \end{subfigure}
    \hfill
    \begin{subfigure}[b]{0.45\textwidth}
        \centering
        \includegraphics[width=.9\textwidth]{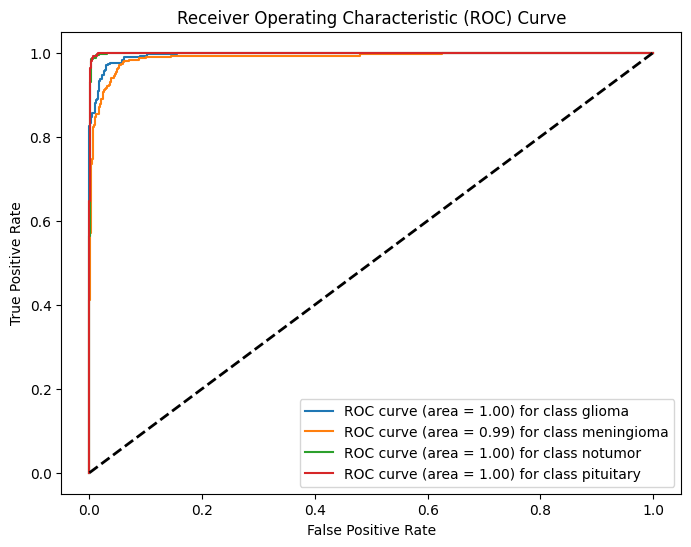}
        \caption{FedProx}
    \end{subfigure}
    \caption{ROC curves for Brain Tumor dataset. Number of Collaborators = 5}\label{fig_bt_ROC}
\end{figure*}
Figure \ref{fig_bt_ROC} shows the ROC curves for both aggregation schemes, FedAvg (left chart) and FedProx (right chart). Each chart shows a different curve for each class of the dataset. The AUC values are reported as well in the legend, for each class. It is possible to appreciate how both schemes are able to obtain very good performance.

\begin{figure}[h]
    \centering
    \includegraphics[width=.9\linewidth]{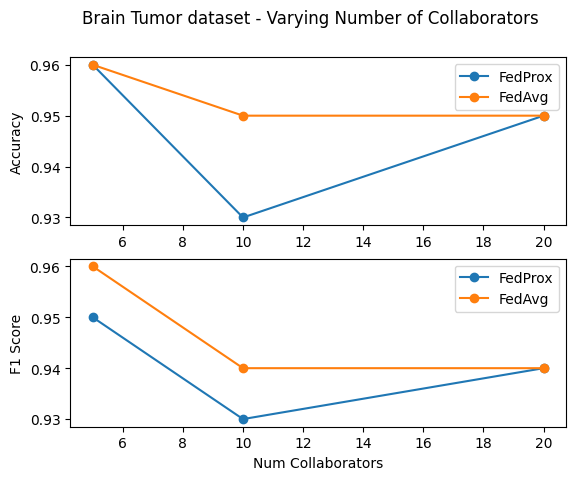}
    \caption{Accuracy and F1 score performances with a varying number of Collaborators}\label{fig:brain_num_nodes}
\end{figure}
Figure\ref{fig:brain_num_nodes} shows the accuracy and F1 score, for both the aggregation methods, when different numbers of Collaborators are used. We can notice that FedAvg remains mostly stable, while FedProx has a decrement when the number of collaborators is equal to 10. This is a result that needs some further evaluation, but that is probably due to the limited size of the dataset. In fact, when we increase the number of Collaborators, the dataset is split into multiple parts and, evidently, not all Collaborators are able to improve the overall performance.

\begin{figure}[h]
    \centering
    \includegraphics[width=.9\linewidth]{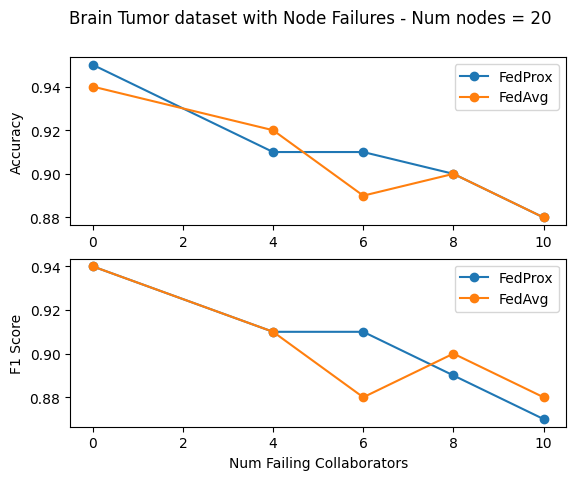}
    \caption{Accuracy and F1 score performances with node failures}\label{fig:brain_failing}
\end{figure}
Figure \ref{fig:brain_failing} reports both the accuracy and F1 score as a function of the number of failing nodes, when the total amount of involved nodes (active and inactive) was equal to 20 Collaborators (plus the Manager). Both aggregation methods, FedProx and FedAvg, are reported. Accuracy and F1 scores of each method change as the number of failing nodes increases.
In general, the trend is that, as expected, as the number of failing nodes increases, the performance decreases. This is due to the fact that part of the whole dataset was distributed among nodes (also failed ones) and the loss of the contribution of certain collaborators affects the training.
It is worth pointing out that we obtained similar results for other settings with different number of nodes in the system (not shown here for the sake of space limits).

Overall, results show that for this dataset the two compared approaches behave quite similarly in terms of performance, hence confirming the effectiveness of using FL in the healthcare sector. They highlight the influence of node failures and the importance of being able to transparently assess the correct execution of the protocol.

\subsection{Federated Learning Smart Contract Gas Consumption}
In this section, we show the gas consumption analysis for the smart contract that governs the interactions in the FL system. Due to space limitations, we will show only results related to system settings with no node failures. 

Table \ref{tab:collab_fee} shows the gas fee consumption for the \emph{send\_weights\_hash()} , i.e., the only non-view function in FLSC called by Collaborators. Results show a low level of dispersion, since the standard deviation is relatively small compared to the mean. This outcome was indeed expected, since with this function a Collaborator registers in the smart contract a hash of the weights it just uploaded to IPFS. Thus, it basically sends a fixed length datum.

\begin{table}
\centering
\caption{Collaborators fee statistics for \emph{send\_weights\_hash()} function}
\label{tab:collab_fee}
\begin{tabular}{|l|r|}
\hline
 & \textbf{gas consumed} \\
\hline
\textbf{mean} & 1390385.00 \\
\textbf{std} & 88532.56 \\
\textbf{median} & 1376425.00 \\
\textbf{min} & 1285740.00 \\
\textbf{25\%} & 1331082.50 \\
\textbf{50\%} & 1376425.00 \\
\textbf{75\%} & 1426507.50 \\
\textbf{max} & 1559830.00 \\
\hline
\end{tabular}
\end{table}


\begin{figure}[h]
    \centering
    \includegraphics[width=.9\linewidth]{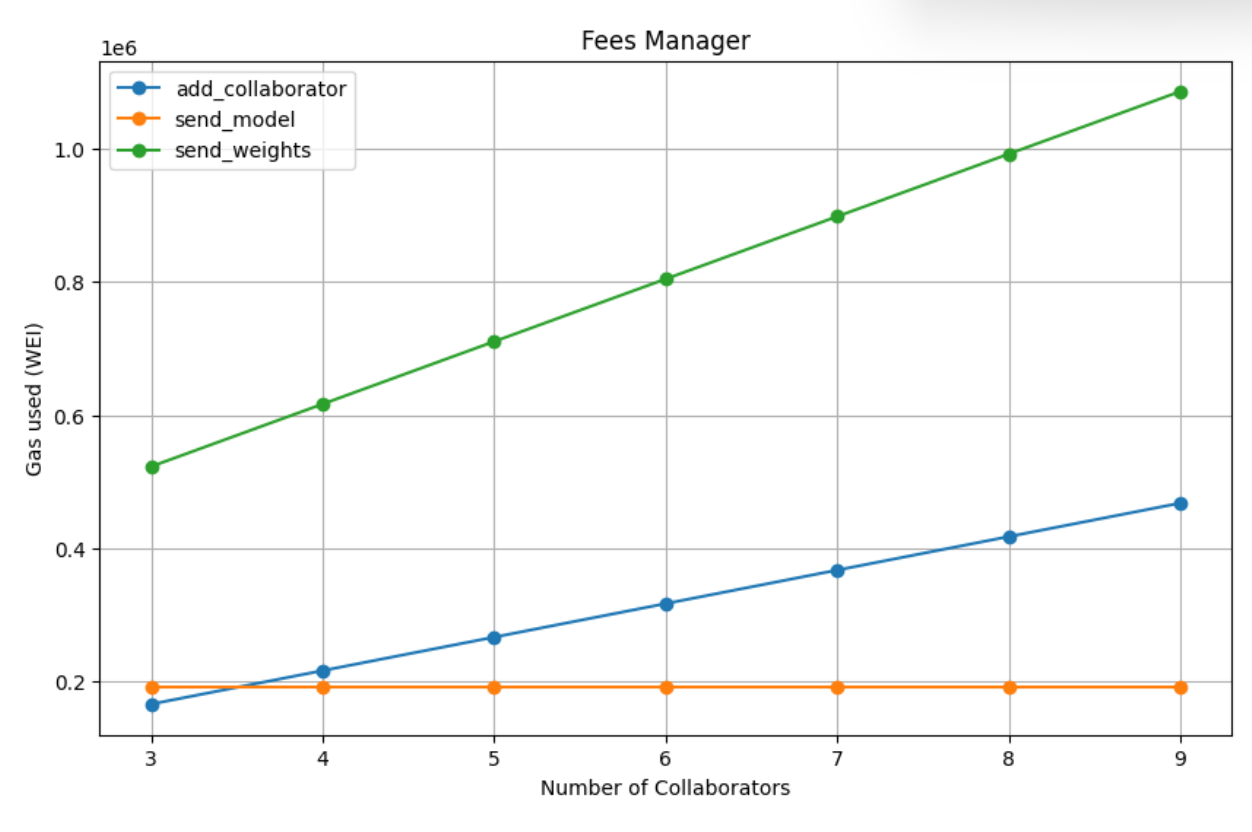}
    \caption{Manager: Gas Fees Incurred}\label{fig:manager_fee}
\end{figure}

Figure \ref{fig:manager_fee} shows gas fees for different methods called by the Manager as a function of the number of Collaborators involved in the FL system. 
Each line represents a different method. 
From the chart, it is possible to observe how the gas fees increase linearly with the amount of nodes in the system, with the exception of \emph{send\_model}, that clearly does not depend on this variable. The cost for the deployment of the contract was $2151147$ gas.
These results confirm the viability of the approach, especially if we assume that the number of Collaborators involved in a FL system is limited and does not require continuous updates to the set of participants. In this latter scenario, maybe some optimization techniques might be necessary.

\subsection{IPFS Delays}
\begin{figure}[h]
    \centering
    \includegraphics[width=.9\linewidth]{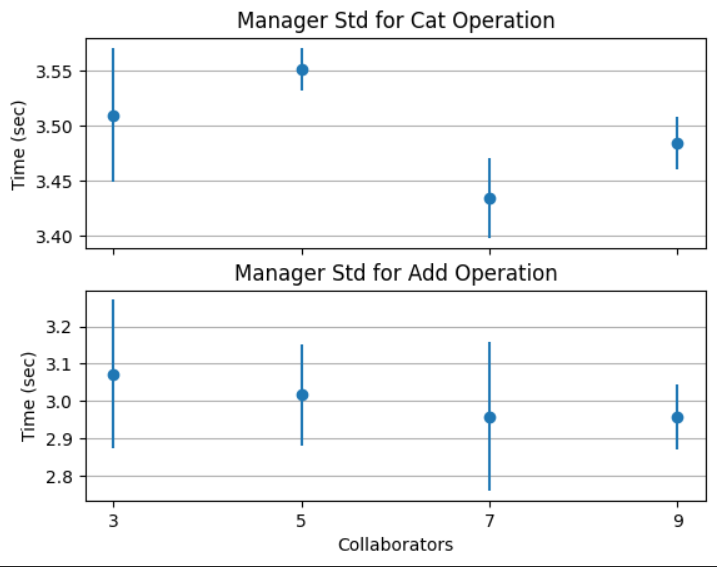}
    \caption{Manager: IPFS delays based on the number of Collaborators}\label{fig:manager_ipfs}
\end{figure}

\begin{figure}[h]
    \centering
    \includegraphics[width=.9\linewidth]{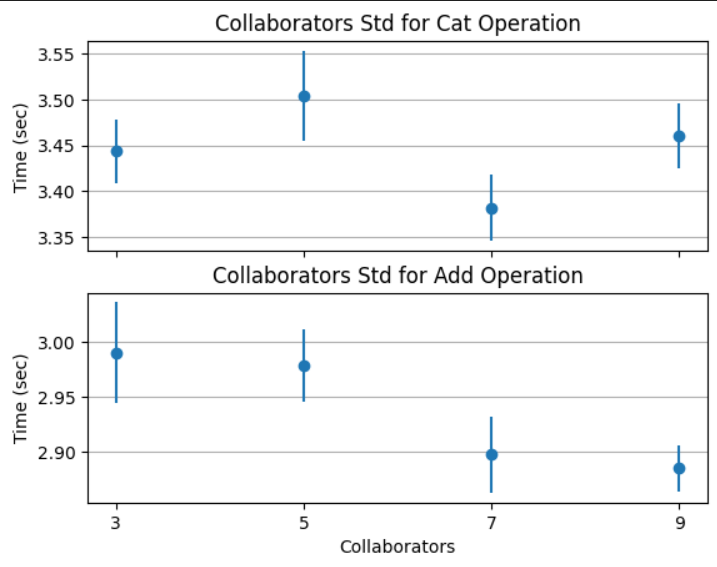}
    \caption{Collaborators: IPFS delays based on the number of Collaborators}\label{fig:collaborators_ipfs}
\end{figure}

Figures \ref{fig:manager_ipfs} and \ref{fig:collaborators_ipfs} show the average times required for the Manager and the Collaborators respectively, to perform essential actions related to updating weights on the Inter-Planetary File System (IPFS). Specifically, we examine the time taken to send (referred to as “Add Operation”) and retrieve (referred to as “Cat Operation”) the updated weights.
The figures display both the averages and standard deviations of the measurements collected. Each individual configuration was repeated 10 times. In each run, the total number of participating nodes was the sum of the Manager and the Collaborators, the number of which is indicated in the chart. The execution spanned 10 rounds of the FL protocol.
In this context, since the metrics measured were related to the transmission of weights to IPFS, all nodes were run on a single server with the following technical specifications: Intel(R) Core(TM) i7-8565U CPU (1.80 GHz - 1.99 GHz), 16.0 GB RAM, and a symmetrical bandwidth connection of 433/433 Mbps. 

Figures show that, as expected, delays are comparable for both the Manager and Collaborators. This consistency arises since the amount of data to be sent or received is the same, i.e., the data comprises a set of numerical values representing the weights of the employed Convolutional Neural Network (CNN). Moreover, the time to retrieve the novel weights was slightly higher than sending them. Finally, average times do not significantly vary based on the amount of involved Collaborators, since the amount of data to be retrieved for each Collaborator remains constant. There are differences in average values and standard deviations, that are emphasized in the charts only because of the limited scale of the y-axis. But in the end, results do not show significant differences.

In summary, however, our measurements confirm that there is a non-negligible time overhead associated with coordinating FL activities.
Indeed, at the end of each round every Collaborator has to upload its weights to IPFS, compute and send the related hash value to the smart contract, and wait for an smart contract event, so as to retrieve the hash from the smart contract and the weights from IPFS. In the middle of this round step, the Manager has to retrieve all the hashes related to each Collaborator from the smart contract, retrieve all the weights of all the Collaborators, compute the average and upload the updated weights on IPFS. Given the measured times, depending on the number of Collaborators this upload phase requires some tens of seconds.

It is important to notice that we observed consistent performance on IPFS even in the event of node failures. Clearly enough, this process remains independent of other nodes, relying solely on the interaction with IPFS.
For this reason and for the sake of brevity, we thus omit these results.

\section{Conclusions}\label{sec:conc}
In this study, we presented a Federated Learning (FL) system that is based on the use of blockchain technology, aiming for a secure and coordinated approach to data management. The design of the framework is centered around data privacy protection, where it exchanges encrypted model parameters instead of sensitive data. This system integrates the Inter-Planetary File System (IPFS) and smart contracts to create a secure, unalterable data storage infrastructure, thereby enhancing data security during FL operations. 
The viability of our proposal is confirmed by the results from our validation and tests. 

The complete code for the project can be accessed here: https://github.com/LorenzoCassano/Blockchain-FederatedLearning/tree/main.

\section*{Acknowledgements}
This work has been partially funded by the European Union - NextGenerationEU within the framework of PNRR Mission 4 - Component 2 - Investment 1.1 under the Italian Ministry of University and Research (MUR) programme "PRIN 2022" - grant number 2022N2NH42 - SmartShires - CUP: H53D23003570006

\bibliographystyle{IEEEtran}
\bibliography{biblio}

\end{document}